\documentclass[conference]{IEEEtran}
\IEEEoverridecommandlockouts
\usepackage{cite}
\usepackage{amsmath,amssymb,amsfonts}
\usepackage{algorithmic}
\usepackage{graphicx}
\usepackage{textcomp}
\usepackage{xcolor}
\usepackage{url}
\usepackage{multirow}
\usepackage{graphicx}
\usepackage[most]{tcolorbox}
\usepackage{titlesec}
\usepackage{listings}
\tcbset{boxrule=0.5pt, arc=2pt, colback=gray!3, colframe=gray!60}
\lstdefinelanguage{json}{
  basicstyle=\ttfamily\small,
  showstringspaces=false,
  breaklines=true,
  columns=fullflexible,
  keepspaces=true,
  morestring=[b]",
  stringstyle=\color{cyan!80!black},
  comment=[l]{//},
  morecomment=[s]{/*}{*/},
  commentstyle=\color{purple!80!black}\itshape,
  literate=
   *{true}{{{\color{green!80!black}true}}}{4}
    {false}{{{\color{red!80!black}false}}}{5}
    {null}{{{\color{orange!80!black}null}}}{4}
}

\newtcblisting{jsonbox}{
  listing only,
  breakable,
  enhanced,
  width=\linewidth,
  colback=white, 
  colframe=gray!40, 
  left=0.8ex, right=0.8ex, top=0.6ex, bottom=0.6ex,
  boxsep=0pt,
  listing engine=listings,
  listing options={
    language=json 
  }
}
\def\BibTeX{{\rm B\kern-.05em{\sc i\kern-.025em b}\kern-.08em
    T\kern-.1667em\lower.7ex\hbox{E}\kern-.125emX}}
\begin{document}

\title{Text2Mem: A Unified Memory Operation Language \\ for Memory Operating System}

\IEEEoverridecommandlockouts
\author{
\IEEEauthorblockN{
Yi Wang\IEEEauthorrefmark{1},
Lihai Yang\IEEEauthorrefmark{1},
Boyu Chen\IEEEauthorrefmark{1},
Gongyi Zou\IEEEauthorrefmark{2},
Kerun Xu\IEEEauthorrefmark{3},\\
Bo Tang\IEEEauthorrefmark{1},
Feiyu Xiong\IEEEauthorrefmark{1},
Siheng Chen\IEEEauthorrefmark{4},
Zhiyu Li\IEEEauthorrefmark{1}\textsuperscript{*}
}
\IEEEauthorblockA{
\IEEEauthorrefmark{1}\textit{MemTensor (Shanghai) Technology}, Shanghai, China \\
\IEEEauthorrefmark{2}\textit{University of Oxford}, Oxford, United Kingdom \\
\IEEEauthorrefmark{3}\textit{National University of Singapore}, Singapore \\
\IEEEauthorrefmark{4}\textit{Shanghai Jiao Tong University}, Shanghai, China
}
\IEEEauthorblockA{
Emails: leo6kwang@gmail.com, lihai@protonmail.com, chenby@memtensor.cn, \\
tangb@memtensor.cn, xiongfy@memtensor.cn, lizy@memtensor.cn, \\
gongyi.zou@oriel.ox.ac.uk, e1521287@u.nus.edu, sihengc@sjtu.edu.cn \\
\textsuperscript{*}Corresponding author
}
}

\maketitle

\begin{abstract}
Large language model agents increasingly depend on memory to sustain long horizon interaction, but existing frameworks remain limited. Most expose only a few basic primitives such as encode, retrieve, and delete, while higher order operations like merge, promote, demote, split, lock, and expire are missing or inconsistently supported. Moreover, there is no formal and executable specification for memory commands, leaving scope and lifecycle rules implicit and causing unpredictable behavior across systems.
We introduce Text2Mem, a unified memory operation language that provides a standardized pathway from natural language to reliable execution. Text2Mem defines a compact yet expressive operation set aligned with encoding, storage, and retrieval. Each instruction is represented as a JSON based schema instance with required fields and semantic invariants, which a parser transforms into typed operation objects with normalized parameters. A validator ensures correctness before execution, while adapters map typed objects either to a SQL prototype backend or to real memory frameworks. Model based services such as embeddings or summarization are integrated when required. All results are returned through a unified execution contract.
This design ensures safety, determinism, and portability across heterogeneous backends. We also outline Text2Mem Bench, a planned benchmark that separates schema generation from backend execution to enable systematic evaluation. Together, these components establish the first standardized foundation for memory control in agents.
\end{abstract}


\section{Introduction}

Large language model (LLM) agents\cite{Zhao2025LLMSurvey,Wang2024AgentSurvey,Luo2025LLMAgentSurvey} are rapidly evolving from single-turn dialogue systems toward long-horizon agents capable of multi-session interaction and extended task execution. In this transition, memory becomes a central capability: it maintains consistent identity, accumulates user preferences, and provides contextual grounding across time, which together enable persistent reasoning and personalized behavior \cite{Yang2024Memory3,Wei2025SecondMe,li2025memos}.  

Yet current memory subsystems remain rudimentary. Most frameworks expose only a small set of basic primitives such as encode, retrieve, and delete. Higher-order controls that are crucial for realistic use, including merge, promote or demote, split, lock, and expire, are either absent or only inconsistently implemented through ad-hoc extensions \cite{Packer2024MemGPT,Chhikara2025Mem0}. This incompleteness creates several obstacles in practice. Portability across systems is limited because the same intent must be redefined for each framework. Compositional task design is difficult because some verbs overlap while others are missing entirely. More fundamentally, everyday use is constrained when users cannot express essential operations for organizing, protecting, or managing the lifecycle of their memories.  

A second obstacle is the lack of a formal and executable specification for memory operations. Natural language commands such as “get rid of old notes” or “make rent top priority” are inherently underspecified: the scope, the mode of deletion, and the governance rules remain implicit. Without a schema that enforces required fields and invariants, and without typed objects that normalize values such as time ranges and priorities, systems cannot determine execution reliably. From the user’s perspective, this means memory commands may behave unpredictably, vary across platforms, or even fail silently. From the developer’s perspective, the absence of a shared specification makes it impossible to design consistent behaviors or to ensure that extensions will interoperate safely.  

\begin{figure*}[t]
    \centering
    \includegraphics[width=0.90\linewidth]{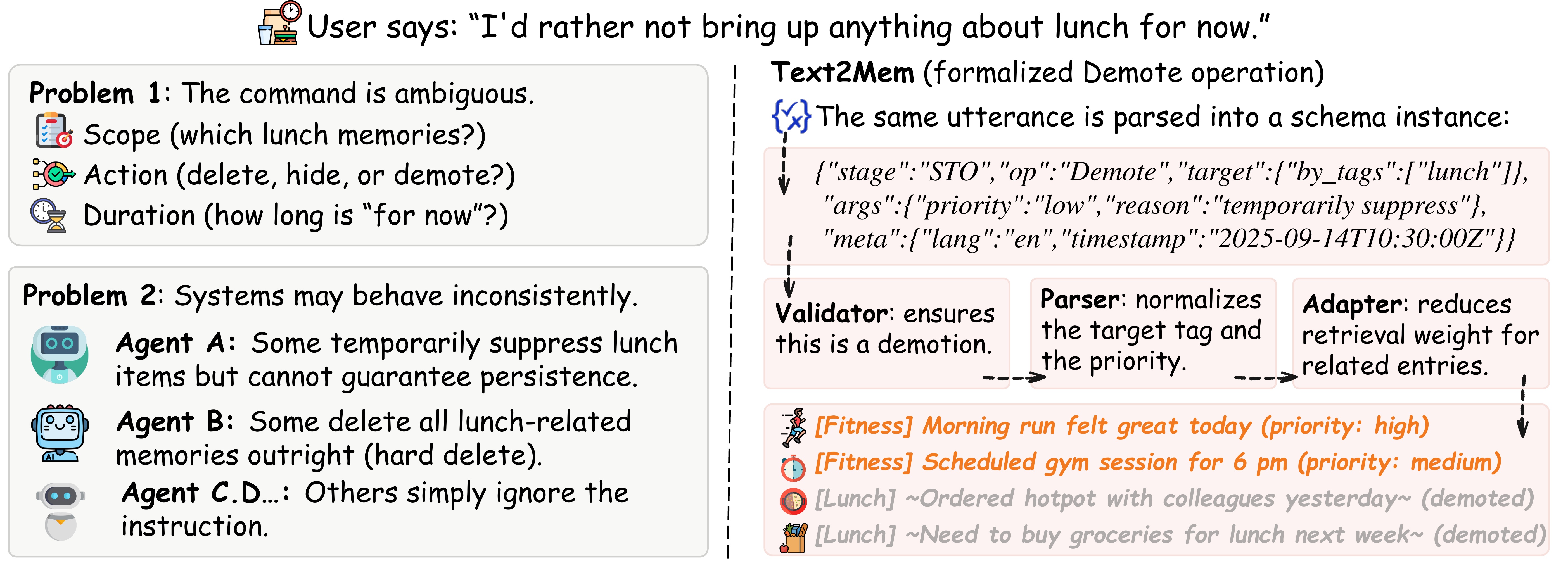}
    \caption{Ambiguity and inconsistency in current systems versus Text2Mem’s formalized handling. 
    \textbf{Left}: The natural language instruction ``I'd rather not bring up anything about lunch for now'' is underspecified. Scope, action, and duration are unclear, leading to inconsistent behaviors across agents (temporary suppression, hard deletion, or ignoring). 
    \textbf{Right}: Text2Mem resolves the ambiguity by instantiating a schema-based Demote operation with explicit arguments. The validator, parser, and adapter guarantee consistent execution across heterogeneous backends.}
    \label{fig:problem}
\end{figure*}

As an illustration, consider the utterance: \emph{``I'd rather not bring up anything about lunch for now.''} 
In existing frameworks, this instruction is ambiguous: it is unclear which memories fall under “lunch,” whether the intent is to delete, hide, or demote them, and how long “for now” should last. 
Different systems therefore behave inconsistently: some temporarily suppress lunch-related items without persistence, others delete them outright, and many simply ignore the command. 
In contrast, Text2Mem formalizes the request as a Demote operation with explicit fields: the target is all memories tagged with lunch, and the args specify a lowered priority rather than deletion. 
The validator ensures that the action is a demotion, the parser normalizes the scope and priority, and the adapter consistently reduces retrieval weight across backends. 
This guarantees that everyday expressions map to predictable and safe memory operations (Figure~\ref{fig:problem}).

This paper introduces Text2Mem, a unified memory operation language that addresses the fragmentation of existing systems. Text2Mem provides a compact but expressive operation set aligned with the cognitive stages of encoding, storage, and retrieval. The language eliminates redundancy while elevating advanced controls such as merge, split, promote or demote, lock, and expire to first-class status with precise semantics. Each operation is expressed through a schema-based specification that enforces required fields and invariants, while a type parser produces strongly typed operation objects ready for execution. Together, these features ensure that memory commands are formally defined, automatically validated, and safely instantiated.  

Beyond language design, Text2Mem enhances portability, consistency, and research reproducibility. The same typed object can be executed in both a SQL-like reference backend and adapters for real frameworks, ensuring consistent behavior across systems. By separating language understanding from backend execution, Text2Mem provides a stable interface for agents, reduces ambiguity in everyday use, and enables replicable studies of long-horizon memory. In sum, Text2Mem establishes the first standardized pathway from natural language to reliable memory control.  

This paper makes the following contributions:  
\begin{itemize}
    \item We propose \textbf{Text2Mem}, the first unified memory operation language for LLM-based agents. It defines a compact but expressive set of twelve operations, spanning encoding, storage, and retrieval, with clear semantic boundaries and support for higher-order controls.  

    \item We introduce a \textbf{schema-based specification} that formally encodes each operation’s required fields, invariants, and constraints, together with a type parser that produces strongly typed operation objects for safe and deterministic execution.  

    \item We demonstrate \textbf{portability across execution backends}: the same typed object can be executed in a SQL-like reference backend and mapped to real memory frameworks, ensuring consistent behavior and enabling reproducible studies of long-horizon memory.  
\end{itemize}

\begin{figure*}[h]
    \centering
    \includegraphics[width=1.0\linewidth]{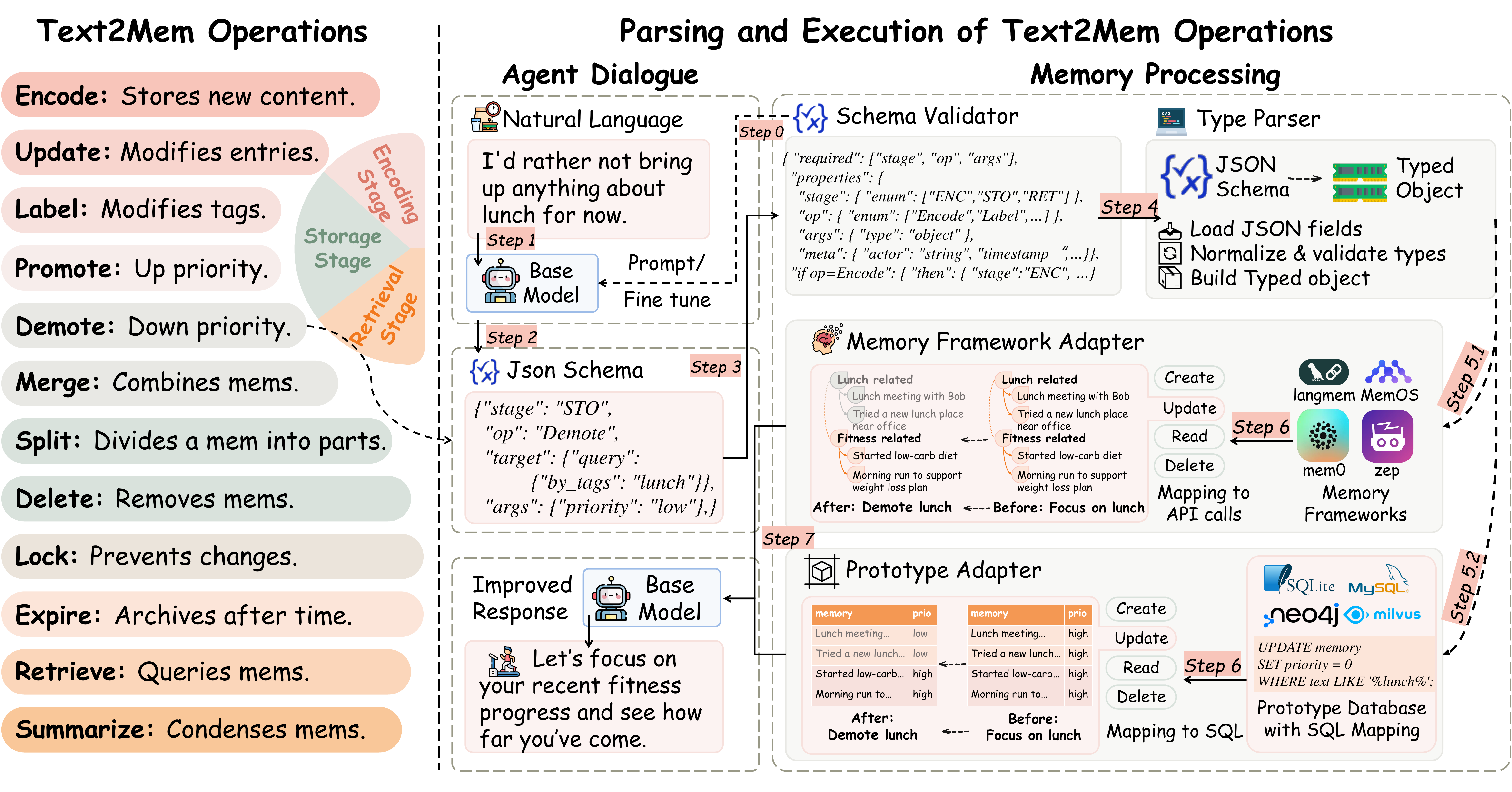}
    \caption{Illustration of the Text2Mem execution pathway. Natural language instructions are normalized into memory operation schema instances, which are validated, parsed into typed operation objects, and finally executed through adapters to real memory frameworks or, alternatively, through a SQL-based prototype backend for controlled verification.}
    \label{fig:text2mem}
\end{figure*}

\section{Background and Motivation}

\subsection{Agent memory and operation frameworks}

Recent work has explored augmenting LLM agents with human-inspired or tool-based memory mechanisms. Early studies draw on theories of human memory, such as hippocampal indexing, to integrate language models with structured stores for more effective consolidation and retrieval \cite{Gutierrez2024HippoRAG,Gutierrez2025FromRAGtoMemory}. Others make transient context explicit, treating attention caches or hierarchical stores as first-class carriers of working memory, thereby reducing inference costs while maintaining recall \cite{Yang2024Memory3}. More functionally oriented systems mimic human behaviors such as note-taking or summarization to improve organization and durability \cite{Liang2024PGRAG,Wei2025SecondMe,Wu2023AutoGen}.  

In parallel, tool-based approaches provide explicit interfaces for editing or extending memory. Parameter-level methods expose APIs for inserting, modifying, or deleting knowledge within models \cite{Zhang2024EasyEdit,Xu2025EasyEdit2}. External memory modules mitigate context-window bottlenecks through extract–update workflows or graph-structured representations \cite{Chhikara2025Mem0,Zhong2024MemoryBank,Rasmussen2025Zep}.  

Building on these foundations, system-oriented designs aim to treat memory as a first-class operating component. \textsc{MemGPT} proposes modularizing context into dynamic pages \cite{Packer2024MemGPT}, \textsc{A-MEM} introduces agentic memory abstractions for LLM agents \cite{Xu2025AMEM}, and \textsc{MemOS} presents a more comprehensive operating system view with richer primitives and scheduling capabilities \cite{li2025memos}. These frameworks make memory manipulation more explicit and practically useful, but they largely remain fragmented. They often expose CRUD-style utilities or ad-hoc extensions, and lack a systematic, semantically precise operation language to unify higher-order controls such as prioritization, consolidation, or lifecycle governance.  

\subsection{Lessons from text-to-SQL}

An instructive parallel comes from research on text-to-SQL. This field has faced the same fundamental challenge as memory-centric agents: natural language is underspecified, ambiguous, and highly variable, yet systems must translate it into precise, executable operations.  

On the modeling side, early systems such as Seq2SQL \cite{Zhong2017Seq2SQL} treated the task as generating SQL queries from utterances. Later work introduced richer architectures, for example RAT-SQL \cite{Wang2020RATSQL} with relation-aware schema encoding, and UniSAr \cite{dou2022UniSAr} with structure-aware generation that integrates schema and language signals. This evolution marks a shift from surface mappings to models that explicitly encode constraints and structural dependencies.  

On the benchmark side, the breakthrough was to introduce a constrained schema layer for standardized evaluation. Spider emphasized cross-domain generalization under a unified schema representation \cite{Yu2018Spider}, and CoSQL extended this idea to conversational settings with clarification and repair grounded in the same schema \cite{Yu2019CoSQL}. More recent efforts such as BIRD and LogicCat highlight that larger scale and deeper reasoning only increase the importance of a shared schema backbone \cite{li2024bird,liu2025LogicCat}.  

This perspective informs our design. Memory commands today are in the same position SQL queries once were: vague in everyday use, inconsistent across implementations, and lacking a common schema. Text2Mem applies the lessons of text-to-SQL by defining a compact but expressive operation set, enforcing schema-level constraints, and grounding execution in typed objects to make memory control precise and reproducible.  

\section{Text2Mem}
\label{sec:text2mem} 

To address the limitations of current systems, we introduce Text2Mem, a memory operation language that provides a standardized pathway from natural language instructions to reliable execution. As illustrated in Figure~\ref{fig:text2mem}, Text2Mem follows a three-stage pipeline. First, free-form utterances are mapped into memory operation schema instances, expressed in a JSON-based format with explicit fields and constraints. Second, a validator checks structural and semantic legality, after which a parser converts schema instances into typed operation objects that normalize arguments such as time ranges, priorities, and tags. Third, an adapter compiles typed objects into concrete actions that can be executed either on a SQL-like prototype database or on the APIs of real memory frameworks. This separation ensures that the same instruction can be translated, verified, and executed consistently across different backends.

The remainder of this section elaborates the core components of Text2Mem: the design of its verb-centered operation set (Section~\ref{subsec:operations}), the schema specification and typed object model (Section~\ref{subsec:schema}), and the validator–parser–adapter pipeline that grounds execution across diverse backends (Section~\ref{subsec:pipeline}). 

\subsection{Operation Set Design}
\label{subsec:operations}

\subsubsection{\textbf{Design Principles}}
\label{subsubsec:ops-principles}

A central component of Text2Mem is its verb-centered operation inventory, which defines how agents communicate memory intentions in a precise and executable way.  

The design of Text2Mem follows three foundational principles that ensure clarity, coverage, and composability in memory operations.  

\paragraph{Mutual Exclusivity.}  
Each verb represents a unique atomic behavior with non-overlapping semantics.  
By enforcing exclusivity, the language avoids ambiguity between similar intents, so that every operation can be independently verified and safely executed.  

\paragraph{Completeness.}
The inventory spans the full memory lifecycle—encoding, storage, and retrieval—within a single coherent vocabulary and schema. 
It supports everyday edits, prioritization, structural transformation, lifecycle and access governance, as well as retrieval and summarization, so that essential agent memory behaviors are expressible without gaps or overlap.

\paragraph{Minimality.}  
Redundant verbs and aliases are eliminated, so that each operation contributes distinct expressive power while remaining composable with others.  
This minimal design prevents conceptual bloat and allows higher workflows to be constructed as clear combinations of atomic operations.  

Together, these three principles establish a unified and auditable language layer for agent memory.  
Unlike ad-hoc APIs or loosely defined commands, Text2Mem provides a stable semantic backbone where every action has a deterministic meaning, explicit boundaries, and interoperable behavior across systems.  
This principled design not only guarantees internal consistency but also enables long-term scalability: new verbs can be safely added without breaking existing semantics, and different backends can interpret the same command with predictable results.  

\subsubsection{\textbf{Memory Operation Set}}
\label{subsubsec:memory-ops}

The final inventory contains twelve verbs: one for \emph{encoding}, nine for \emph{storage}, and two for \emph{retrieval} (Table~\ref{tab:ops_compare}). Each verb has a distinct functional scope; together they cover the full lifecycle of agent memory without semantic overlap.

\paragraph{Encoding.}

The encoding stage defines how new information becomes memory rather than raw data. 

In Text2Mem, a memory item is typed, addressable, and semantically grounded. It carries content, context such as who, when, and where, facets such as tags, entities, and topics, governance fields including priority, permissions, lineage, and lifespan, and machine-usable signals such as embeddings, keywords, and summaries. 

The single verb \textbf{Encode} creates such an item through semantic interpretation rather than a store-and-embed shortcut. It resolves entities and time, normalizes expressions, extracts structure such as tasks, decisions, and references, and attaches provenance and salience before any indexing or embedding is produced. 

A request like remember the key points from today’s meeting becomes a structured record with decisions, dated action items, responsible owners, and links to materials; embeddings are added afterwards for retrieval. Saving a paper as a reference yields citation metadata such as title, authors, venue, and year, topical entities, and an access policy. A vacation photo is encoded with place and time, people, and event tags. 

In practice, \textbf{Encode} first clarifies what the item is and how it should live in memory. It resolves entities and time, attaches source, author, and audience, extracts decisions and tasks when present, sets initial governance such as priority, permissions, and lifespan, and chooses an appropriate granularity. Only then does it produce lexical indices, embeddings, and concise summaries that conform to the schema. 

By treating encoding as understanding before indexing, Text2Mem records each item with clear meaning, scope, and provenance. The result is a well-formed, typed unit that is easy to interpret, govern, and reuse, and that remains consistent across different backends. 

\begin{table*}[h]
\renewcommand{\arraystretch}{1.2}
\centering
\footnotesize
\caption{The Text2Mem operation inventory and its support in existing frameworks. 
\checkmark: native support; $\triangle$: partial or indirect; --: unsupported. 
Basic operations like \texttt{Encode}, \texttt{Delete}, and \texttt{Retrieve} are common, 
while higher-order controls remain absent.}
\begin{tabular}{l|l|l|c|c|c}
\hline
\textbf{Stage} & \textbf{Operation} & \textbf{Description} & \textbf{MemOS} & \textbf{mem0} & \textbf{Letta} \\
\hline
Encoding 
  & \texttt{Encode} 
  & Insert a new memory with metadata; embeddings are optional and deferrable later. 
  & \checkmark & \checkmark & \checkmark \\
\hline
\multirow{9}{*}{Storage} 
  & \texttt{Update}   
  & Modify specific fields with validation, lineage safety, and strict type checks. 
  & \checkmark & $\triangle$ & $\triangle$ \\
  & \texttt{Label}    
  & Add, replace, or remove tags and edit facets with deduplication constraints. 
  & $\triangle$ & $\triangle$ & $\triangle$ \\
  & \texttt{Promote}  
  & Increase priority or attach reminders to resurface items on a defined cadence. 
  & -- & -- & -- \\
  & \texttt{Demote}   
  & Decrease priority or archive without deleting data to reduce retrieval prominence. 
  & -- & -- & -- \\
  & \texttt{Merge}    
  & Combine records into a primary while preserving lineage links and optional child deletes. 
  & -- & -- & -- \\
  & \texttt{Delete}   
  & Soft or hard delete with policy and lock checks, including time-range filters. 
  & \checkmark & \checkmark & \checkmark \\
  & \texttt{Split}    
  & Break composite entries into smaller linked units via sentences or chunks. 
  & -- & -- & -- \\
  & \texttt{Lock}     
  & Restrict edits via read-only or append-only policies with reason and expiry metadata. 
  & -- & -- & -- \\
  & \texttt{Expire}   
  & Apply TTL or until, triggering actions when expired such as demote or anonymize. 
  & -- & -- & -- \\
\hline
\multirow{2}{*}{Retrieval} 
  & \texttt{Retrieve} 
  & Run filtered, ranked queries with permission-aware results and field-level whitelists. 
  & \checkmark & \checkmark & \checkmark \\
  & \texttt{Summarize}
  & Produce concise, focused summaries within token limits aligned to a specified focus. 
  & $\triangle$ & $\triangle$ & $\triangle$ \\
\hline
\end{tabular}
\label{tab:ops_compare}
\end{table*}

\paragraph{Storage.}

The storage stage defines how memories evolve after they are created. 
Unlike a traditional database that centers on CRUD over opaque rows, Text2Mem treats storage as semantic governance over well-formed memory items. 
Each operation can invoke language-model capabilities to interpret, rewrite, and consolidate content, while policy fields maintain priority, permissions, lineage, and lifespan.

\textbf{Update} is not a blind field overwrite. It can use generation to improve clarity, correct inconsistencies, normalize entities and time, or expand a terse note into an actionable task with owner and deadline. 
\textbf{Label} goes beyond manual tagging by inferring topics and entities from content and context, aligning them with existing taxonomies. 
\textbf{Delete} supports governance-aware removal, including soft deletion with recovery windows and hard deletion subject to locks and audit trails. 

\textbf{Promote} and \textbf{Demote} introduce priority as a first-class control for memory salience. They influence ranking, recency decay, and reminder cadence, enabling the agent to foreground objectives or quiet background chatter without rewriting content. 
These controls are essential in practice yet largely absent from prior memory frameworks; Text2Mem makes them part of the core vocabulary. 

\textbf{Merge} and \textbf{Split} operate at semantic granularity. 
With vector retrieval, the system can gather related items and use the model to consolidate duplicates or near-duplicates into a coherent entry while preserving provenance links. 
Conversely, lengthy or multi-topic notes can be split into atomic pieces that are easier to retrieve and govern, with explicit lineage between parent and children. 

\textbf{Lock} and \textbf{Expire} provide safety and lifecycle guarantees. 
Locks can enforce read-only or append-only modes for sensitive items, while expiration can attach a time-to-live or a concrete date with on-expire behavior such as archival or demotion. 
Both operations are audit-friendly and interact with other fields without ambiguity. 

Taken together, the nine storage verbs—\textbf{Update}, \textbf{Label}, \textbf{Delete}, \textbf{Promote}, \textbf{Demote}, \textbf{Merge}, \textbf{Split}, \textbf{Lock}, \textbf{Expire}—lift storage from data maintenance to memory governance. 
They enable rewriting with understanding, consolidation with lineage, prioritization without mutation, and lifecycle control with predictable outcomes, which are difficult to realize in traditional databases. 

\paragraph{Retrieval.}
The retrieval stage brings memory back into focus for reasoning and reuse. 

\textbf{Retrieve} issues filtered and ranked queries using a hybrid strategy that combines symbolic predicates with embedding similarity. 
It respects governance fields such as permissions, priority, soft deletes, and expiration, and can apply freshness or diversity controls to avoid stale or redundant results. 
Typical prompts include find recent research discussions, surface open action items, or retrieve notes about client A across meetings. 

\textbf{Summarize} performs semantic condensation over one or many retrieved items. 
Instead of truncation, it produces concise narratives or structured digests—decisions, action items with owners and dates, open questions—optionally paired with a vector summary for later reuse. 
This supports long-horizon continuity by retaining what matters while keeping the working context compact. 

Together, \textbf{Retrieve} and \textbf{Summarize} turn the store into a controllable memory interface: precise when filtering, semantic when ranking, and succinct when carrying knowledge forward. 

\subsubsection{\textbf{Implications and Comparison}}
\label{subsubsec:ops-implications}

After defining the twelve operations across encoding, storage, and retrieval, it is clear that Text2Mem is more than a list of verbs.  
It functions as a coherent language layer that links human intent, symbolic representation, and executable control through unified semantics and validation.  

\paragraph{Design implications.}
Each Text2Mem verb is a first-class semantic primitive that can be composed into larger workflows—for instance, “Retrieve → Label → Promote → Summarize”—while preserving interpretability and determinism.  
The inventory thus provides not only expressive coverage but also a structured grammar for memory reasoning, enabling agents to construct, govern, and reflect on their own memory processes in a consistent manner.  

\paragraph{Comparison to existing frameworks.}
Table~\ref{tab:ops_compare} compares Text2Mem with representative frameworks (MemOS~\cite{li2025memos}, mem0~\cite{Chhikara2025Mem0}, and Letta~\cite{Packer2024MemGPT}).  
While common primitives such as \textbf{Encode}, \textbf{Delete}, and \textbf{Retrieve} are widely supported, higher-order operations—\textbf{Promote}, \textbf{Demote}, \textbf{Merge}, \textbf{Split}, \textbf{Lock}, and \textbf{Expire}—are largely missing or inconsistently implemented.  
This fragmentation underscores the need for a unified specification like Text2Mem, which provides completeness, semantic clarity, and consistent execution across heterogeneous backends.

\subsection{Operation Schema}
\label{subsec:schema}

A central question in memory system design is how to turn natural language instructions—often vague, incomplete, or underspecified—into reliable, executable operations.  
The \textbf{operation schema} of Text2Mem addresses this challenge by defining a minimal yet expressive contract.  
Each command is represented as a typed JSON object with explicit fields and built-in constraints, ensuring that memory control is interpretable, verifiable, and portable across heterogeneous backends.

\subsubsection{\textbf{Design Principles}}

The operation schema is built on three guiding principles that ensure clarity, parsimony, and reliability in memory control.

\textbf{Explicitness.}
Every aspect that affects execution—scope, time, or permissions—must be stated explicitly rather than inferred.  
By making all assumptions visible, the schema eliminates ambiguity and guarantees consistent interpretation across systems.

\textbf{Minimality.}
All operations share a compact backbone of five fields: \textbf{stage}, \textbf{op}, \textbf{target}, \textbf{args}, and \textbf{meta}.  
This minimal structure is expressive enough to cover encoding, storage, and retrieval while avoiding redundancy, making it suitable for both atomic actions and multi-step workflows.

\textbf{Safety.}
Validation precedes execution.  
Potentially destructive actions require confirmation or dry-run modes, and lifecycle constraints prevent invalid operations such as deleting locked or expired items.  
These built-in checks shift safety from implementation to schema design, ensuring deterministic and auditable behavior.

\subsubsection{\textbf{Schema Architecture}}

Each operation in Text2Mem is represented as a typed, executable contract following a compact yet expressive backbone of five fields: \textbf{stage}, \textbf{op}, \textbf{target}, \textbf{args}, and \textbf{meta}.  
\textbf{Stage} defines the cognitive layer—encoding (\textbf{ENC}), storage (\textbf{STO}), or retrieval (\textbf{RET})—and is automatically inferred from the operation verb, enabling deterministic routing within the execution pipeline.  
\textbf{Op} specifies one of twelve mutually exclusive verbs that capture the system’s full memory lifecycle, ensuring that each action has unambiguous semantics.  
\textbf{Target} identifies the scope of affected memory items, acting as the boundary between user intent and system execution.  
\textbf{Args} contains the verb-specific parameters that modify or query those targets, while \textbf{Meta} records contextual information such as the actor, timestamp, language, and safety switches (\textbf{confirmation} or \textbf{dry\_run}).  
Together, these five fields form a self-describing and auditable object that can be validated, logged, and executed uniformly across heterogeneous backends.  
This structure enables schema instances to serve not merely as API calls but as verifiable execution contracts—portable across systems yet semantically stable within any runtime.

\textbf{Target} defines scope and is intentionally minimal yet auditable—choose exactly one: (i) \textbf{ids} (explicit identifiers), (ii) \textbf{filter} (structured predicates over type/tags/time/weight), (iii) \textbf{search} (semantic or keyword intent), or (iv) \textbf{all} (global scope, exceptional only).  
Filters and searches in storage must include a numeric limit; using \textbf{all=true} requires \textbf{confirmation} or \textbf{dry\_run} in \textbf{meta}; retrieval with \textbf{all} must confirm explicitly.  
This design makes range control predictable: narrow scopes execute directly, wide writes demand confirmation, and global actions are opt-in.

\textbf{Core constraints} are minimal but strict: \textbf{Encode} requires a \textbf{payload}; \textbf{Label} requires tags or facets (with mode add/replace/remove); \textbf{Update} requires a \textbf{set} object; \textbf{Promote}/\textbf{Demote} use mutually exclusive \textbf{weight} or \textbf{weight\_delta} (plus reminder/archive options); \textbf{Expire} needs a finite horizon (\textbf{ttl} or \textbf{until}) with an explicit post-action; \textbf{Lock} specifies mode (read\_only/append\_only).  
Cross-field invariants enforce governance—locked items cannot be hard-deleted, expired items cannot be updated, and numeric/temporal values are clamped to valid ranges—so safety is guaranteed by schema, not by ad-hoc procedural code.

\subsubsection{\textbf{Illustrative Workflows}}

Beyond atomic operations, Text2Mem supports multi-step workflows where each stage is independently validated yet semantically coherent.  
The following two examples demonstrate how the schema generalizes from targeted control to system-level governance.

\paragraph{Example 1: Semantic Promotion Workflow.}
This example illustrates a lightweight, retrieval-driven update.  
A project owner wants to raise the importance of all OKR-related notes while leaving unrelated entries unchanged.  
The workflow encodes two notes and then promotes those matched through semantic search.

\textbf{Step 1: Encode relevant and background notes.}
\begin{jsonbox}
{"stage":"ENC","op":"Encode",
 "args":{"payload":{"text":"Key task: refine OKR review mechanism"},
         "tags":["OKR","priority"],
         "type":"note",
         "time":"2025-04-12T16:00:00+08:00",
         "source":"miro_board",
         "facets":{"subject":"OKR review","topic":"action item"}}}
\end{jsonbox}

\begin{jsonbox}
{"stage":"ENC","op":"Encode",
 "args":{"payload":{"text":"Meeting note: OKR metrics progress plan"},
         "tags":["meeting","OKR"],
         "type":"note",
         "time":"2025-04-13T09:00:00+08:00",
         "source":"feishu_doc",
         "facets":{"subject":"weekly meeting","topic":"metrics update"}}}
\end{jsonbox}

These commands insert two memory items with consistent facets and tags, allowing later semantic linkage.

\textbf{Step 2: Promote via semantic search.}
\begin{jsonbox}
{"stage":"STO","op":"Promote",
 "target":{"search":{"intent":{"query":"OKR"},
                     "overrides":{"limit":3},
                     "limit":3}},
 "args":{"weight":0.9}}
\end{jsonbox}

The search-based target bridges semantic retrieval and memory governance.  
Only items matched by vector or lexical similarity are affected, and the explicit \textbf{limit} parameter guarantees bounded scope.  
This workflow shows how Text2Mem supports intent-based, explainable updates that remain deterministic across backends.

\paragraph{Example 2: Incident Postmortem Archive.}
The second example demonstrates a complex multi-stage workflow that integrates encoding, governance, and retrieval.  
After a SEV-1 network outage, an SRE engineer records a war-room timeline, locks the records for compliance, and generates an executive summary for leadership review.

\textbf{Step 1: Encode incident timeline.}
\begin{jsonbox}
{"stage":"ENC","op":"Encode",
 "args":{"payload":{"text":"2025-09-28 API P1 incident timeline: 20:07 alert, 20:12 failover, 20:28 routing misconfig, 20:41 fix released, 20:48 95% recovered."},
         "tags":["incident:p1-network","postmortem","owner:sre-ling"],
         "type":"war_room_timeline",
         "time":"2025-09-28T22:30:00+08:00",
         "facets":{"subject":"2025-09-28 API Outage","topic":"incident_response"},
         "location":"cn-shanghai"}}
\end{jsonbox}

This step persists the event record as structured memory, tagged for later retrieval and governance.

\textbf{Step 2: Lock incident records for review.}
\begin{jsonbox}
{"stage":"STO","op":"Lock",
 "target":{"filter":{"has_tags":["incident:p1-network"],
                     "time_range":{"start":"2025-09-28T00:00:00+08:00",
                                   "end":"2025-10-05T23:59:59+08:00"},
                     "limit":200}},
 "args":{"mode":"read_only",
         "reason":"Preserve SEV-1 timeline for legal and executive review",
         "policy":{"allow":["Retrieve","Summarize"],
                   "deny":["Update","Delete"],
                   "reviewers":["oncall_manager","sre_lead"],
                   "expires":"2025-12-31T23:59:59+08:00"}},
 "meta":{"actor":"sre-ling","timestamp":"2025-09-29T00:05:00+08:00"}}
\end{jsonbox}

The lock command freezes the affected entries as read-only.  
Policy fields define who can access or summarize them and when the restriction expires, embedding governance directly into schema-level constraints.

\textbf{Step 3: Generate postmortem summary.}
\begin{jsonbox}
{"stage":"RET","op":"Summarize",
 "target":{"search":{"intent":{"query":"2025-09-28 API outage follow-up",
                               "context":"executive briefing"},
                     "overrides":{"k":8,"order_by":"time_desc"},
                     "limit":8}},
 "args":{"focus":"Highlight root cause, customer impact, and assigned remediation owners."},
 "meta":{"actor":"cto-office","lang":"en"}}
\end{jsonbox}

The final step retrieves relevant entries and synthesizes an English executive summary.  
Together, these steps form a fully auditable memory workflow—typed, validated, and safe to execute in any backend.

\textbf{Summary.}
The two examples demonstrate complementary facets of Text2Mem’s schema:  
the first emphasizes semantic selectivity and controlled prioritization, while the second showcases lifecycle governance and cross-role coordination.  
In both cases, natural-language intentions are decomposed into deterministic, verifiable operation sequences, proving the schema’s expressiveness and robustness for real-world memory management.

\subsubsection{\textbf{Discussion and Implications}}
The Text2Mem schema functions not merely as a data format but as a governance layer for memory control.  
Its compact design encodes safety, interpretability, and determinism directly into structure, transforming natural language commands into verifiable execution plans.  
This layer makes memory operations portable across systems while preserving auditability, forming the formal foundation of the Text2Mem language.

\subsection{Validator--Parser--Adapter Pipeline}
\label{subsec:pipeline}

\subsubsection{\textbf{Pipeline Overview}}
Once natural language instructions are normalized into schema instances, Text2Mem executes them through a three-layered pipeline that guarantees structural validity, semantic determinism, and backend consistency.  
The \textbf{validator} enforces schema correctness and rejects unsafe or underspecified commands; the \textbf{parser} converts validated instances into typed operation objects with normalized parameters; and the \textbf{adapter} maps those objects into concrete backend actions through either SQL-based simulation or real memory frameworks.  
All stages return a unified \textbf{ExecutionResult} containing execution status, affected entries, and resulting state changes.  
This modular design ensures that (1) malformed instructions never reach execution, (2) valid instances are deterministically interpreted, and (3) identical typed objects behave equivalently across heterogeneous systems.

\subsubsection{\textbf{Validator}}
The validator constitutes the first safeguard of the pipeline.  
It verifies every schema instance against the structural and semantic rules defined in the Text2Mem specification.  
At the structural level, it enforces required fields, data types, and enumerated values.  
At the semantic level, it checks cross-field invariants: locked items cannot be hard-deleted, expiration must include a finite horizon, and global writes must be explicitly confirmed.  
When violations are detected, the validator halts execution and returns a structured diagnostic that identifies the failing field and rule.  
By shifting safety checks from runtime to validation, this stage guarantees that all downstream processing begins with well-formed, interpretable instructions.

\subsubsection{\textbf{Parser}}

The parser converts validated schema instances into typed operation objects—the canonical internal form used by Text2Mem.  
Its role is to make every instruction explicit, structured, and machine-determinable.  
Each field is assigned a precise type, and parameters are normalized to eliminate ambiguity and redundancy.  
Implicit relations are expanded into clear key–value pairs, ensuring that all temporal, logical, and access rules are fully specified before execution.  
If inconsistencies or missing references are detected, the parser halts execution and returns a diagnostic error.  
By enforcing a uniform and interpretable internal representation, the parser guarantees that all downstream operations behave predictably and can be audited across systems.

\subsubsection{\textbf{Adapter}}
The adapter bridges typed operation objects to executable actions in real or simulated backends.  
It supports two complementary pathways and integrates optional model-driven services when required.

\textbf{Mapping to real frameworks.}  
For operational memory systems such as MemGPT, mem0, or Letta, the adapter translates each typed operation object into the corresponding framework-specific API sequence.  
Rather than performing a one-to-one parameter mapping, this translation occurs at the semantic level: Text2Mem preserves the intent and governance rules of each operation while adapting to the target framework’s execution model.  

For example, a \textbf{Promote} command with a reminder is decomposed into two sub-operations—priority adjustment and scheduled notification—implemented respectively through the framework’s ranking or scheduling interfaces.  
A \textbf{Lock} command updates access control tables and enforces review policies by configuring the framework’s permission layer.  
Similarly, a \textbf{Merge} request triggers a retrieval step to collect relevant entries and delegates their consolidation to the system’s internal merge API or, when unavailable, to an auxiliary LLM service.  
In retrieval frameworks that support query vectors or hybrid search, a \textbf{Summarize} operation can invoke both database-level selection and model-level compression, returning a standardized summary object to the caller.  

Through this mapping layer, identical typed objects yield equivalent and auditable behavior across heterogeneous frameworks.  
It also decouples schema evolution from backend implementation, allowing Text2Mem to act as a unifying abstraction layer that integrates symbolic operations, memory governance, and model-assisted reasoning under a common execution protocol.

\textbf{Mapping to the SQL prototype.}  
Beyond dedicated memory frameworks, Text2Mem also provides a reference backend that maps operations to a relational database, enabling full execution even in environments without a specialized memory system.  
In this mode, every operation is compiled into a sequence of SQL statements augmented with optional language-model calls when semantic reasoning is required.  

Simple operations follow direct mappings: \textbf{Encode} becomes an \texttt{INSERT} with text, tags, embeddings, and governance metadata; \textbf{Update} and \textbf{Delete} compile into \texttt{UPDATE} statements that modify or deactivate existing entries while preserving lineage.  
More complex verbs are executed as hybrid symbolic–semantic workflows.  
For instance, \textbf{Merge} first performs a retrieval query to collect candidate records, then invokes an LLM to generate a merged summary, which is inserted as a new item linked bidirectionally to its sources.  
\textbf{Promote} adjusts numeric weights or reminders through SQL field updates, while optionally logging the reason or source context for auditability.  
\textbf{Summarize} issues a \texttt{SELECT} constrained by target filters, retrieves matching items, and calls a summarization model to produce an aggregated textual report that is stored as a derived memory.  
This SQL pathway serves as a transparent and auditable reference implementation: it traces every intermediate step, supports hybrid execution with model assistance, and provides deterministic logs for verification, benchmarking, and downstream agent integration.

\textbf{LLM integration.}  
Certain operations rely on language model capabilities.  
\textbf{Encode} may invoke embedding services for semantic retrieval, and \textbf{Summarize} calls summarization or abstraction models.  
These derived outputs—embeddings, summaries, or keywords—are then reattached to the memory backend through the same unified interface, maintaining schema-level traceability.

\subsubsection{\textbf{Summary}}
The validator, parser, and adapter form a continuous, fault-tolerant execution path from natural language to concrete action.  
Validation ensures safety, parsing ensures normalization, and adaptation ensures consistency across environments.  
Together they establish the operational backbone of Text2Mem—turning underspecified user intent into deterministic, auditable, and portable memory operations.

\begin{figure*}[h]
    \centering
    \includegraphics[width=1.0\linewidth]{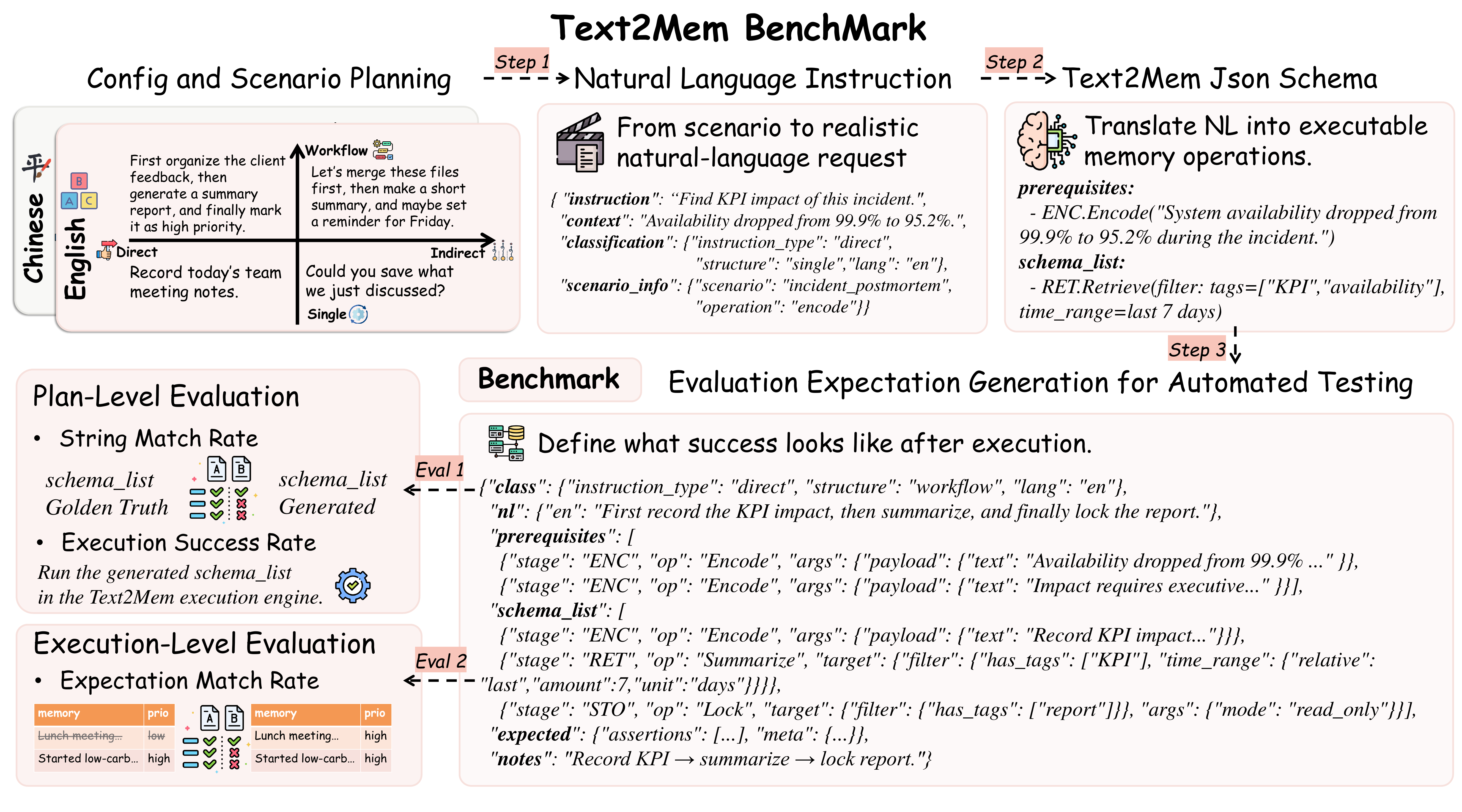}
    \caption{
    \textbf{Overview of the Text2Mem Benchmark pipeline.}
    The process consists of three generation stages and two evaluation layers.
    \textbf{Step~1:} Scenarios are converted into realistic natural-language requests.
    \textbf{Step~2:} The requests are translated into executable memory operation schemas (\texttt{schema\_list}) with corresponding \texttt{prerequisites}.
    \textbf{Step~3:} Expected outcomes are defined for automatic verification after execution.
    Two evaluation layers assess system performance:
    \textbf{Plan-level evaluation} measures string-match accuracy and execution success rate of generated schemas;
    \textbf{Execution-level evaluation} measures expectation match rate and retrieval-based metrics to quantify behavioral correctness after running the \texttt{schema\_list} in the Text2Mem system.}
    \label{fig:benchmark}
\end{figure*}

\section{Text2Mem Bench}

Text2Mem Bench provides an end-to-end benchmark for assessing memory-centric reasoning systems.  
It evaluates not only how models understand and formalize memory-related instructions, but also how faithfully these formalized operations execute within a dynamic memory environment.  
By coupling natural-language interpretation with verifiable database effects, the benchmark bridges intent understanding and operational reliability under a unified, auditable framework.

\subsection{Evaluation Methodology}
Text2Mem Bench evaluates both the \emph{planning layer}—how accurately a system translates natural language into valid Text2Mem schemas—and the \emph{execution layer}—how reliably those schemas produce verifiable effects within a memory environment.  
To enable reproducible and interpretable benchmarking, all evaluations are conducted inside a structured SQL-based prototype that mirrors real memory frameworks while remaining fully auditable.

\paragraph{Structured memory environment.}
We construct a unified memory database that captures the full semantics of Text2Mem operations.  
Each record is a typed, addressable item with fields for content, semantic facets, importance, embeddings, provenance, lifecycle metadata, locks, lineage, and permissions.  
This schema supports the entire memory lifecycle—encoding, storage, and retrieval—while ensuring every operation leaves a persistent and queryable footprint for evaluation.

\paragraph{Scenario construction and context generation.}
At the beginning of each test, a blank memory database is instantiated to ensure a clean and reproducible environment.  
The scenario then executes a sequence of predefined setup operations—such as inserting notes, tasks, events, and references—to construct a semantically coherent context.  
This generated context serves as the interpretive background for the benchmark instruction, enabling the system to retrieve, relate, or modify prior information in a controlled and realistic manner.  
By initializing the environment procedurally rather than statically, the benchmark ensures both reproducibility and fidelity to real-world memory dynamics.

\paragraph{Planning (Natural Language $\rightarrow$ Text2Mem Schema).}
Given a natural-language instruction, the system must generate a \emph{memory operation schema} that satisfies three criteria:  
(i) it is structurally valid JSON under the Text2Mem specification;  
(ii) it is semantically and syntactically well-formed—fields, slots, and parameters are correctly typed and directly normalizable by the parser into \emph{typed operation objects} such as time ranges, priorities, and targets; and  
(iii) for composite or multi-step commands, it produces a coherent sequence of operations (\emph{schema\_list}) that preserves execution order and dependency across stages.  
Together, these criteria evaluate how effectively a system can translate natural-language intent into executable and compositional memory control instructions.

\paragraph{Execution (Text2Mem Schema $\rightarrow$ Real Effects).}
Given a validated schema instance (or an ordered \emph{schema\_list}), the system must not only conform to the formal grammar and semantics of Text2Mem, but also realize the intended \emph{operational effects} in real execution environments.  
This stage bridges the gap between specification and action—analogous to verifying that syntactically valid pseudocode can be executed as functioning code.  
The benchmark therefore evaluates whether each schema leads to consistent, measurable outcomes across both the SQL-based reference backend and real framework adapters.  
Expected effects include state transitions for editing operations, ranked outputs for retrieval, and temporal triggers such as reminders or expirations.  
All results are \emph{database-verifiable} through declarative assertions that track content updates, priority adjustments, and lifecycle transitions, ensuring that formally correct schemas also produce functionally correct behavior.

\begin{table*}[t]
\renewcommand{\arraystretch}{1.12}
\setlength{\tabcolsep}{4.8pt}
\centering
\footnotesize
\caption{Structured expectation templates for the twelve Text2Mem operations. 
Each operation is defined by its pre-state, post-state, and verification expression, 
enabling automated evaluation through SQL assertions.}
\begin{tabular}{l|l|l|l|l}
\hline
\textbf{Stage} & \textbf{Op} & \textbf{Pre-state} & \textbf{Post-state} & \textbf{Check expression} \\
\hline
Encoding & Encode & ID not in DB & New record with content and metadata & $\Delta$count = +1 \\ 
\hline
\multirow{9}{*}{Storage} 
& Update & Record exists & Fields updated; lineage preserved & val\_aft=exp $\wedge$ lid\_aft=lid\_bef \\
& Label & Tags/facets exist & Tag set modified, deduped & DISTINCT(tags) $\wedge$ tags\_aft $\neq$ tags\_bef \\
& Promote & weight=w$_0$ & weight increased or trigger added & weight\_aft $>$ weight\_bef $\vee$ EXISTS(trigger) \\
& Demote & weight=w$_0$ & weight decreased; record active & weight\_aft $<$ weight\_bef \\
& Merge & Multi-ID src & Children merged; lineage linked & merged\_into $\neq$ NULL $\wedge$ count(child)=1 \\
& Delete & Record active & Flagged deleted or removed & $\Delta$count = -n \\
& Split & Composite record & Child records linked to parent & count(child)$>$1 $\wedge$ pid=src \\
& Lock & Editable record & Lock set (RO/AO) & lock $\in$ \{RO,AO\} \\
& Expire & TTL unset & Expiry registered; trigger active & expiry $\neq$ NULL $\wedge$ EXISTS(trigger) \\
\hline
\multirow{2}{*}{Retrieval} 
& Retrieve & Query-matching records & Results ranked and filtered & ids=exp $\wedge$ rank=exp \\
& Summarize & Context records & Summary stored with refs & sim $\geq$ $\tau$ \\
\hline
\end{tabular}
\label{tab:expectations}
\end{table*}

\paragraph{Metrics and reproducibility.}
The benchmark reports metrics independently for the planning and execution layers, each defined through formally verifiable quantities.

At the \emph{planning layer}, we measure two core metrics:  
\begin{itemize}
    \item \textbf{String Match Accuracy (SMA):} the proportion of generated schemas that exactly match their gold references.  
    \item \textbf{Execution Success Rate (ESR):} the proportion of schemas that can be parsed and executed without error in the benchmark engine.  
\end{itemize}

Let $\mathcal{S}$ denote the set of generated schemas, $\mathcal{S}^*$ the corresponding gold references,  
and $\psi(\cdot)$ an execution validator returning 1 if a schema runs successfully.  
The two metrics are defined as
\[
\mathrm{SMA} =
\frac{1}{|\mathcal{S}|}
\sum_{s \in \mathcal{S}} 
\mathbf{1}[s = \mathcal{S}^*],
\]
\[
\mathrm{ESR} =
\frac{1}{|\mathcal{S}|}
\sum_{s \in \mathcal{S}}
\psi(s).
\]
SMA evaluates exact structural and semantic fidelity, whereas ESR captures robustness and executability under real parsing and normalization.

At the \emph{execution layer}, performance is quantified by the \textbf{Expectation Match Rate (EMR)}, 
the ratio of satisfied assertions to all verifiable conditions.  
Each operation type defines a structured template—pre-state, post-state, and verification expression—instantiated as SQL-level assertions during evaluation.  
Table~\ref{tab:expectations} summarizes these templates across the twelve operations.

Let $\mathcal{A}_i$ denote the set of assertions for instance $i$, and $\mathbf{1}[\cdot]$ the indicator of satisfaction.  
The EMR is computed as
\[
\mathrm{EMR} =
\frac{\sum_i \sum_{a \in \mathcal{A}_i} \mathbf{1}[a]}
{\sum_i |\mathcal{A}_i|}.
\]
Typical checks include record creation ($\Delta n{>}0$ for Encode), weight increase ($w_1{>}w_0$ for Promote), and tag modification ($t_1{\neq}t_0$ for Label).  
Together, SMA, ESR, and EMR provide a unified framework for assessing syntactic accuracy, operational reliability, and execution correctness across the full Text2Mem pipeline.

\subsection{Dataset Construction}
We organize benchmark instances along a four-way taxonomy that controls linguistic difficulty, operational structure, language, and evaluation layer.  
This taxonomy ensures balanced coverage of realistic memory-use scenarios while keeping the benchmark compact, interpretable, and reproducible.  
Each instance is automatically validated under the Text2Mem schema and linked to executable SQL assertions, guaranteeing verifiable transitions from instruction to effect.

\subsubsection{\textbf{Scenario Planning.}}
Each scenario defines a minimal yet semantically rich context that anchors instruction interpretation.  
Instances are derived from realistic sources such as meeting notes, task trackers, and project documentation, ensuring that every memory operation is tested under plausible work and knowledge-management situations.

\paragraph{Instruction type: direct vs.\ indirect.}
\emph{Direct} instructions explicitly express both the intended operation and its parameters.  
For example, “raise the priority of note 42 to high and set a reminder for tomorrow” directly corresponds to a promotion operation with a defined priority and time trigger.  
Such cases primarily test structural accuracy and schema formatting.

\emph{Indirect} instructions, in contrast, encode intent implicitly through pragmatic cues or conversational context.  
They require the system to infer missing arguments and determine the appropriate operation.  
For instance, “stop bringing up the lunch topic for a while” implies a demotion or temporary expiration, while “can you make this easier to find later?” entails a promotion through tagging or resurfacing.  
This dimension therefore measures a model’s ability to resolve underspecified or context-dependent language into explicit, executable memory operations.

\paragraph{Structure: single vs.\ workflow.}
\emph{Single} instructions correspond to a single atomic operation whose outcome can be evaluated independently, such as adding a note, updating a tag, or deleting an obsolete record.  
These instances isolate the correctness of individual schema generation and execution.

\emph{Workflow} instructions, by contrast, compose several dependent operations that share intermediate state—typically three to five sequential steps.  
A typical workflow might first retrieve a subset of meeting notes, then label them as “urgent,” promote key items, and finally summarize them into a concise report.  
The benchmark runner executes these steps transactionally, preserving inter-step dependencies and verifying final outcomes through aggregated assertions.  
This dimension tests a system’s ability to plan coherent multi-operation sequences, maintain consistency, and propagate references across stages.

\paragraph{Language: English vs.\ Chinese.}
Each instance is available in English (nl.en) or Chinese (nl.zh), and a subset provides bilingual pairs.  
Monolingual items evaluate schema grounding within a single language, while bilingual ones support cross-lingual analysis of planning fidelity and schema consistency.

\subsubsection{\textbf{Dataset Generation Process.}}
The construction of Text2Mem Bench follows a three-stage pipeline that mirrors the planning–execution workflow of the benchmark itself.  
Each stage transforms natural data into structured, executable, and verifiable instances.

\paragraph{Stage I: Context synthesis.}
Raw materials are collected from realistic work and knowledge traces such as meeting minutes, task logs, and collaborative notes.  
A minimal but semantically rich context is synthesized for each test case, containing heterogeneous items (notes, tasks, events, and references).  
This contextual grounding provides the environmental state against which subsequent instructions are interpreted.

\paragraph{Stage II: Schema generation.}
Natural-language instructions—either direct or indirect, single or workflow—are paired with corresponding Text2Mem schemas through a semi-automatic annotation pipeline.  
Automatic templates are first generated using high-precision schema synthesis rules and model-assisted alignment, followed by human verification.  
Each finalized schema conforms to the official JSON specification and is independently validated through structural parsing.

\paragraph{Stage III: Assertion binding.}
For every schema, a set of declarative SQL assertions is automatically instantiated based on the operation type (e.g., insert, update, merge, retrieve).  
These assertions specify expected database state transitions, providing executable ground truth for evaluation at the execution layer.  
The final dataset therefore contains end-to-end verifiable samples—from instruction to schema to post-execution effects—ensuring reproducibility and precise measurement across both layers.

\paragraph{Final Verification and Benchmark Compilation.}
Once all schemas and assertions have been validated for correctness, the fully verified instances are retained as the official benchmark dataset.  
This curated set ensures that all test cases are reproducible, accurate, and ready for consistent evaluation across models and frameworks.  
The benchmark is then packaged for public release, providing a robust and executable set of examples for future research and development.

\paragraph{Benchmark Statistics and Upcoming Results.}
The benchmark dataset is composed of a diverse range of test cases, ensuring comprehensive coverage of both simple and complex scenarios across all dimensions of the taxonomy.  
Each instance is annotated with metadata, including difficulty level, operation type, and language pair, allowing for detailed statistical analysis.  
In the coming months, we will publish results from a series of tests across various foundational models, evaluating their performance on the Text2Mem benchmark.  
These results will include detailed metrics on planning and execution accuracy, as well as cross-backend consistency, providing valuable insights into the capabilities and limitations of current models in handling structured memory tasks.

\section{Conclusion}
\label{sec:conclusion}

This paper introduced Text2Mem, a unified memory operation language for agents. The design consists of an operation set aligned with encoding, storage, and retrieval, a schema based specification that enforces fields and invariants, typed operation objects for deterministic parsing, and adapters that execute consistently across a prototype database and real frameworks. The result is a standardized pathway from natural language to reliable memory control with clear guarantees of safety, determinism, and portability. A planned benchmark will evaluate the pathway end to end by separating planning from execution and by comparing behavior across backends. We believe this language layer enables more predictable everyday use and provides a stable basis for future evaluation and system integration. 

\bibliographystyle{IEEEtran}
\bibliography{ref}

\end{document}